\documentclass[10pt]{article}
\usepackage{url}
\usepackage{graphics} 
\usepackage{epsfig} 
\usepackage{times} 
\usepackage{amsmath} 
\usepackage{amssymb}  
\usepackage{algorithm}
\usepackage{algorithmic}
\usepackage[caption=false,font=footnotesize]{subfig}
\usepackage{float}
\usepackage{array}
\usepackage{balance}
\usepackage{bm}
\usepackage{txfonts}
\usepackage{epstopdf}
\usepackage{geometry}
\usepackage{boldline}
\usepackage{array}

\newcolumntype{L}[1]{>{\raggedright\let\newline\\\arraybackslash\hspace{0pt}}m{#1}}
\newcolumntype{C}[1]{>{\centering\let\newline\\\arraybackslash\hspace{0pt}}m{#1}}
\newcolumntype{R}[1]{>{\raggedleft\let\newline\\\arraybackslash\hspace{0pt}}m{#1}}

\DeclareMathOperator*{\argmin}{argmin}
\DeclareMathOperator*{\argmax}{argmax}

\title{Towards Recovery of Conditional Vectors from\\Conditional Generative Adversarial Networks}

\author{Sihao Ding\qquad Andreas Wallin \\
\texttt{\{sihao.ding, andreas.wallin1\}@volvocars.com} \\
}
\date{\vspace{-2ex}}

\begin{document}

\maketitle

\begin{abstract}
A conditional Generative Adversarial Network allows for generating samples conditioned on certain external information.
Being able to recover latent and conditional vectors from a conditional GAN can be potentially valuable in various applications, ranging from image manipulation for entertaining purposes to diagnosis of the neural networks for security purposes.
In this work, we show that it is possible to recover both latent and conditional vectors from generated images given the generator of a conditional generative adversarial network.
Such a recovery is not trivial due to the often multi-layered non-linearity of deep neural networks.
Furthermore, the effect of such recovery applied on real natural images are investigated.
We discovered that there exists a gap between the recovery performance on generated and real images, which we believe comes from the difference between generated data distribution and real data distribution.
Experiments are conducted to evaluate the recovered conditional vectors and the reconstructed images from these recovered vectors quantitatively and qualitatively, showing promising results.
\end{abstract}

\section{Introduction}\label{sec:introduction}
A Generative Adversarial Network (GAN)~\cite{goodfellow2014generative} is a generative model that can produce realistic samples from random vectors drawn from a known distribution. A GAN consists of a generator $G$ and a discriminator $D$, both of which are usually implemented as deep neural networks. The training of a GAN involves an adversarial game between the generator and the discriminator. In the context of images, the generator maps low-dimensional vectors from latent space to image space, creating images that are intended to come from the same distribution as the training data; the discriminator tries to classify between images generated by the generator (trying to assign score $0$) and real images from training data (trying to assign score $1$). Ideally, the distribution of the images generated from the generator become indistinguishable from the distribution of real images in the training set, and the discriminator assigns $0.5$ to both generated and real images. In practice, this is hard to achieve and there is usually a gap between the learned distribution by generator and the real distribution.

A conditional Generative Adversarial Network, sometimes called a cGAN~\cite{{mirza2014conditional},{gauthier2014conditional}} is an extension from GAN which allows for generating samples conditioned on certain external information. Such an extension takes the form of feeding the conditional vector into both the generator and the discriminator during the training process. After training, the generator can generate samples dictated by the condition from a random vector together with a conditional vector. The ability to control certain attributes of the generated samples is of crucial importance for a lot of applications, such as image inpainting~\cite{pathak2016context}, image manipulation~\cite{zhu2016generative}, style transfer~\cite{li2016precomputed}, future frame prediction~\cite{mathieu2015deep}, text-to-image~\cite{reed2016generative} and image-to-image translation in general~\cite{{isola2016image},{CycleGAN2017}}.

Recovering the latent vector as well as the conditional vector from an image can be useful. It is known that vectors that are close in latent and conditional space generates visually similar images, and algebraic operations in latent vector space often lead to meaningful corresponding operations in image space~\cite{radford2015unsupervised}. For a given image, being able to access the latent and conditional vector allow us to perform many tasks such as realistic image editing, data augmentation, inferences, retrieval, compression, and other insights of what the networks see and learn which can be significant to debugging, diagnosis, and other security related issues. The original cGAN framework and GAN framework in general do not provide a straightforward way of reverse back from an image to latent and conditional vector. We cover some of the previous work on recovering latent vector of a GAN in Section~\ref{sec:related-work}. In this work, we show that it is also possible to recover the conditional vector from a cGAN for a known generator. While the recovery of latent vectors may become unreliable under the effect of mode collapse~\cite{{che2016mode},{salimans2016improved}} when different latent vectors are mapped to a single image, the recovery of conditional vectors are usually robust, since it is rare for a successfully trained cGAN to map different conditional vectors to the same image.

A very important point to make here is that it is not the same to recover from an image generated by the generator, and from a real image. Recovering the latent and conditional vector from generated image can be considered an reverse operation which the forward operation does exist. However, when recovering from a real image we are treating it as if it was generated by the generator whereas in fact such a mapping may not exist. Thus, it is more like a projection of a real image onto the manifold learned by the generator. Besides recovering from generated images, this more interesting question of whether sensible conditional information can be recovered from real images is investigated in this work.

\section{Related Work}\label{sec:related-work}
It is out of the scope of this work to conduct a comprehensive literature review of GAN, we point the readers to a good summary given in~\cite{goodfellow2016nips}. Next we discuss some closely related work on recovery/inverting from image domain to vector domain.

The problem of recovering input of a deep neural networks is non-trivial due to the non-linearity, multi-layers and high-dimensional space of a deep neural network.
In~\cite{mahendran2015understanding} they proposed to invert a convolutional neural network (CNN) to gain insights of the hidden layers of the network. In~\cite{donahue2016adversarial} and~\cite{dumoulin2016adversarially}, both groups proposed to learn an auxiliary network during the training of GAN in order to map the generated images back to their latent vectors.
In~\cite{zhu2016generative} images are projected back to the manifold learned by the generator by learning a deep network that minimizes loss based on further extracted CNN features, which is suitable for natural scenes. Such methods of utilizing an auxiliary network to map images back to latent space have advantage of fast mapping, however, requires training an extra network during the training of $G$ and $D$ networks, and cannot always achieve robust precision.

A gradient-based approach is proposed by~\cite{creswell2016inverting}. The evaluation is done in image domain using reconstruction loss, with no report of reconstruction of latent vectors. In fact, we find out later in our experiments that it can take much longer for two latent vectors to become almost identical than it takes for their generated images to become visually indistinguishable. Recent work by~\cite{lipton2017precise} proposed to recover latent vector using a gradient-based method with ``stochastic clipping'', and achieve successful recovery $100\%$ of time given a certain residual threshold. The idea of ``stochastic clipping'' is based on the fact that latent vectors are continuous and have close to zero probability of landing on the boundary values, which doesn't always generalize to conditional vectors in a cGAN framework.

Our work is build on~\cite{creswell2016inverting} and~\cite{lipton2017precise}, showing it is possible to recover conditional vector in a cGAN. The recovery process does not involve simultaneously training an auxiliary network coupled with the original cGAN, which makes it more flexible and possible to apply on trained cGANs. Moreover, we examine the effect of such recovery on real images besides generated images, which is less addressed in previous works.

\section{Recovery Approach}\label{sec:recovery-approach}
In a non-conditional GAN setting, the generator takes a latent vector $\bm{z}\in\varmathbb{R}^{d_z}$ from a known distribution (usually uniform or Gaussian) as input and generates a sample $G(\bm{z})\in\varmathbb{R}^{d_I}$. Here $d_z$ and $d_I$ are dimension of the latent vector and image, respectively. To recover $\bm{z}$ from $G(\bm{z})$, a probe vector $\bm{z_p}\in\varmathbb{R}^{d_z}$ is randomly initialized. The goal is to find a $\bm{z_p}$ such that the $G(\bm{z_p})$ generated from it is identical as $G(\bm{z})$. Ideally, this $\bm{z_p}$ will be the recovery of $\bm{z}$. Following~\cite{lipton2017precise}, this process can be formulated as an optimization shown in Eq.~\ref{eq:objective-function-original}.

\begin{equation}\label{eq:objective-function-original}
\bm{z_p^{*}} ~=~ \argmin\limits_{\bm{z_p}} ~~\|G(\bm{z})-G(\bm{z_p})\|_2^2
\end{equation}

This is optimized using a gradient-based method, with a \emph{stochastic clipping} method introduced in~\cite{lipton2017precise}. The idea is to randomly assign a value to $\bm{z_p}[i]$ if the $i^{th}$ dimension of $\bm{z_p}$ is outside of the range of allowed value (e.g., $[-1,~1]$ assuming $\bm{z_p}$ is drawn from a uniform distribution in $[-1,~1]$) during optimization. Another intuition for doing so is that the probability of a randomly drawn value falls right on the boundary is close to zero. This in our opinion is similar to random re-initialization potentially multiple times to get out of impossible value ranges.

Under the conditional GAN setting, a conditional vector $\bm{y}\in\varmathbb{R}^{d_y}$, where $d_y$ is the dimension of conditional vector, is feed into the generator together with the latent vector $\bm{z}$ (Fig.~\ref{fig:cgan}). Following the same logic, now two probe vectors $\bm{z_p}$ and $\bm{y_p}$ are randomly initialized and optimized iteratively so that $G(\bm{z_p},~\bm{y_p})$ approaches $G(\bm{z},~\bm{y})$. Notice that the latent vector and the conditional vector needs to be optimized simultaneously, updating $\bm{y_p}$ without updating $\bm{z_p}$ can lead to incorrect solution. Eq.~\ref{eq:objective-function-original} can be modified for the conditional setting into Eq.~\ref{eq:objective-function-cond} shown below:

\begin{equation}\label{eq:objective-function-cond}
\bm{z_p^{*}},~\bm{y_p^{*}} ~=~ \argmin\limits_{\bm{z_p}, ~\bm{y_p}} ~~\|G(\bm{z},~\bm{y})-G(\bm{z_p},~\bm{y_p})\|_2^2
\end{equation}

This can be optimized using the same approach as above only if $\bm{y}$ also takes continuous value like $\bm{z}$. However, in most cases the conditional vectors takes discrete integer values and are fed into networks in one-hot encoding~\cite{mirza2014conditional}. Here we specifically consider the solution for one-hot encoding for two reasons: firstly, because one can always easily convert conditional vectors that serve as (multi-dimensional) discrete labels into one-hot encoding; and secondly it helps avoid the time consuming branch-and-bound approach in a typical mix integer programming (MIP) problem. To this end, we formulate our optimization problem as Eq.~\ref{eq:objective-function-cond-reg}:

\begin{equation}\label{eq:objective-function-cond-reg}
\bm{z_p^{*}},~\bm{y_p^{*}} ~=~ \argmin\limits_{\bm{z_p}, ~\bm{y_p}} ~~\|G(\bm{z},~\bm{y})-G(\bm{z_p},~\bm{y_p})\|_2^2~+~\lambda\,|\,\|\bm{y_p}\|_1 - \bm{1}|
\end{equation}

We relax the constraint of $\bm{y_p}$ taking only integer values (0 and 1 in one-hot encoding). To still reach the desired one-hot encoding solution, a regularizer is added to the objective function. $\lambda$ is a constant multiplier. The $L_1$ norm is used to pursue sparsity which is the case in one-hot encoding. The absolute difference between $L_1$ norm of $\bm{y_p}$ and $\bm{1}$ is to enforce the $L_1$ norm be as close as 1. The entire function is minimized when $\bm{y_p}$ is exactly one-hot encoded. Later we will see that this regularization while not having significant impact on recovery from generated images, is important for recovery from real images to obtain reliable results. Again, $\bm{z_p}$ and $\bm{y_p}$ should be optimized together, optimizing one without the other may lead to incorrect combination of $\bm{z_p}$ and $\bm{y_p}$. During optimization, the ``stochastic clipping'' is applied to $\bm{z_p}$ after gradient descent, and a ``projected gradient descent'' is applied to $\bm{y_p}$. More specifically, any value less than 0 is mapped to 0, and any value greater than 1 is mapped to 1. In practise, we find it is better to initialize $\bm{y_p}$ as a zero vector instead of a random one-hot vector, so that the algorithm is not initialized with a false prior information. The overall process is detailed in Algorithm~\ref{alg:1}. $\nabla_{\bm{z_p}}L$ and $\nabla_{\bm{y_p}}L$ are the gradients with respect to $\bm{z_p}$ and $\bm{y_p}$. Notice that the final $\bm{y_p}$ will be reported as $\argmax \bm{(y_p)}$, since we know the true $\bm{y}$ is one-hot encoded.

\begin{algorithm}
\caption{Recovering latent and conditional vector from conditional GAN}\label{alg:1}
\begin{algorithmic}
\STATE{\textbf{function} Recover($G(\bm{z})$)}
\STATE{\qquad $\bm{z_p}\sim U(-1,1)\in\varmathbb{R}^{d_z}$}
\STATE{\qquad $\bm{y_p} ~\leftarrow ~\bm{0}\in\varmathbb{R}^{d_y}$}
\STATE{\qquad\textbf{while} not converged~\textbf{do}}
\STATE{\qquad\qquad $L ~\leftarrow ~\|G(\bm{z},~\bm{y})-G(\bm{z_p},~\bm{y_p})\|_2^2~+~\lambda\,|\,\|\bm{y_p}\|_1 - \bm{1}|$}
\STATE{\qquad\qquad $\bm{z_p} ~\leftarrow ~\bm{z_p}-\alpha\,\nabla_{\bm{z_p}}L$}
\STATE{\qquad\qquad $\bm{y_p} ~\leftarrow ~\bm{y_p}-\beta\,\nabla_{\bm{y_p}}L$}
\STATE{\qquad\qquad $\forall~z_p \in\bm{z_p}$ \textbf{if} $z_p<-1$~\textbf{or} $z_p>1$~\textbf{do} $z_p\sim U(-1,1)$}
\STATE{\qquad\qquad $\forall~y_p \in\bm{y_p}$ \textbf{if} $y_p<0$~\textbf{do} $y_p~\leftarrow~0$}
\STATE{\qquad\qquad $\forall~y_p \in\bm{y_p}$ \textbf{if} $y_p>1$~\textbf{do} $y_p~\leftarrow~1$}
\STATE{\qquad\textbf{end while}}
\STATE{\qquad\textbf{return} $\bm{z_p}$, $\argmax(\bm{y_p})$}
\STATE{\textbf{end function}}
\end{algorithmic}
\end{algorithm}

As mentioned in the Introduction Section, it is different to recover from a generated image than from a real image. Since the forward operation for a generated image does exist, one expects that the conditional vector, being a dominant factor towards generated images, can be recovered even using Eq.~\ref{eq:objective-function-cond} without any constraint on $\bm{y_m}$, at least after the $\argmax$ operation. However, for real images, it is highly likely for a generator to unable to generate their identical copies. It is possible that after projecting an real image onto the learned manifold, it falls onto a spot outside of the defined domain of $\bm{y_p}$. In this case, it is important to have the regularizer as in Eq.~\ref{eq:objective-function-cond-reg} so that the real images are projected onto spots that are semantically explainable in conditional domain.

\section{Experiments}\label{sec:experiments}
\subsection{Experiment Setups}\label{sec:experiment-setups}
Experiments are conducted on two public dataset, MNIST~\cite{lecun1998gradient} and CelebA~\cite{liu2015faceattributes}. For MNIST dataset, the cGAN is trained conditioned on digit classes $0, 1, ..., 9$, making $\bm{y}$ a 10-dimensional vector. For CelebA dataset, we picked two attributes from ground-truth as a proof-of-concept, namely \emph{Female/Male} and \emph{WithGlasses/WithoutGlasses}. The combination of these two attributes is converted to one-hot encoding of 4 classes (0: \emph{Female-WithoutGlasses}, 1: \emph{Male-WithoutGlasses}, 2: \emph{Female-WithGlasses}, 3: \emph{Male-WithGlasses}), leads to a 4-dimensional $\bm{y}$ to train the cGAN. $\bm{z}$ has 100 dimensions and is drawn from uniform distribution $U(-1,1)$ for both datasets. For MNIST dataset the images are zero-padded to a resolution of $32\times32$ with single channel, for CelebA dataset the images are center-cropped and resized to a resolution of $64\times64$ with 3 channels. All pixel values are shift and scaled to $[-1,1]$.
\begin{figure*}[th!]
\centering
\includegraphics[width=0.9\columnwidth]{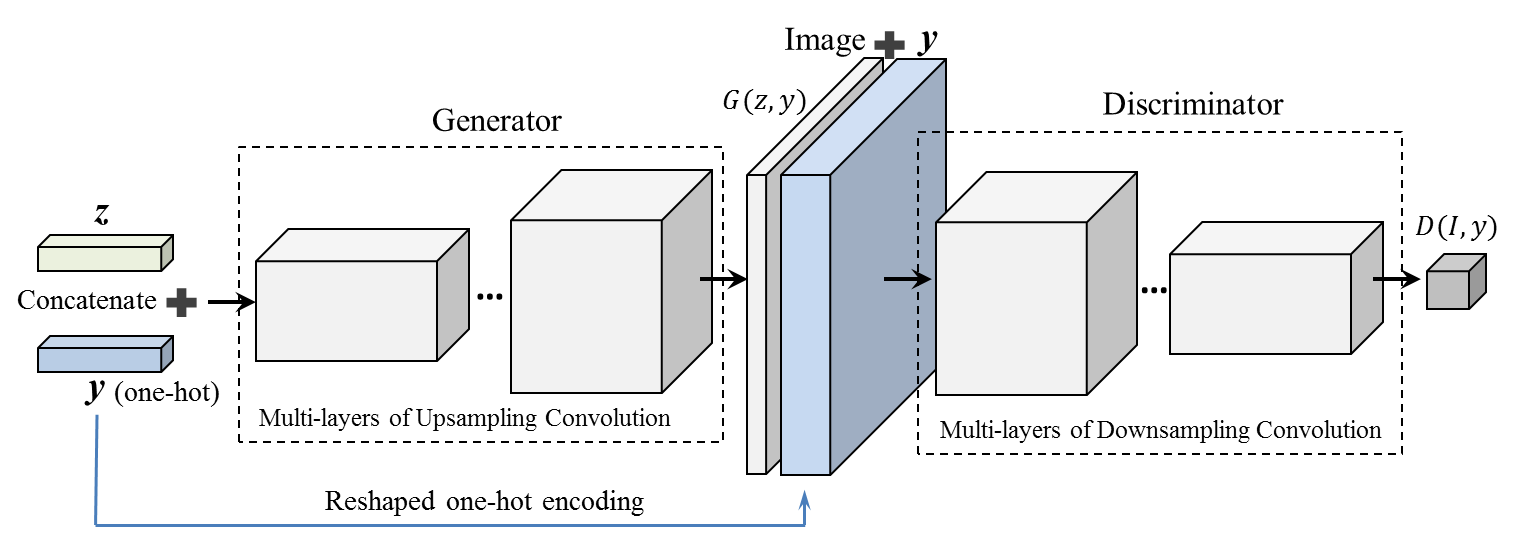}
\caption{The cGAN model used in the experiments. The conditional vector are feeded into both the Generator and the Discriminator using one-hot encoding. It is reshaped (maintaining one-hot encoding) in order to concatenate with noise vector (for Generator) and input image (for Discriminator) along depth channel.}
\label{fig:cgan}
\end{figure*}

The cGAN used in the experiments is a conditional version of DCGAN~\cite{{radford2015unsupervised},{carpedm20}}, as shown in Fig.~\ref{fig:cgan}. The conditional vectors $\bm{y}$ are concatenated with latent vectors $\bm{z}$ as the input for the generator, and are shaped into the image resolution (still in one-hot encoding) and concatenated with generated or real images along depth dimension as the input for the discriminator. No other skip connections are made for the conditional vectors. The discriminator here is the ``vanilla'' binary one instead of a multi-class one that's usually seen in semi-supervised learning. The batch size for MNIST and CelebA datasets are $256$ and $144$, respectively. For recovering, $\lambda=1/d_y$, and $\alpha=\beta=1$ and is reduced to $0.5$ after $50k$ iterations.

\subsection{Recovery from Generated Images}\label{sec:recover-generated-images}
\begin{figure}[h!]
  \centering
  \subfloat[MNIST\label{fig:mnist_g_v}]{\includegraphics[width=0.34\columnwidth]{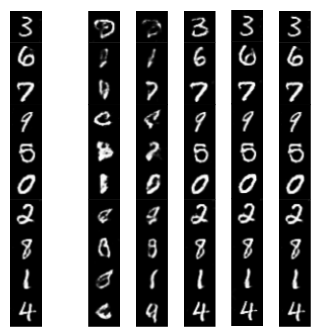}}~~~~
  \subfloat[CelebA\label{fig:celeb_g_v}]{\includegraphics[width=0.64\columnwidth]{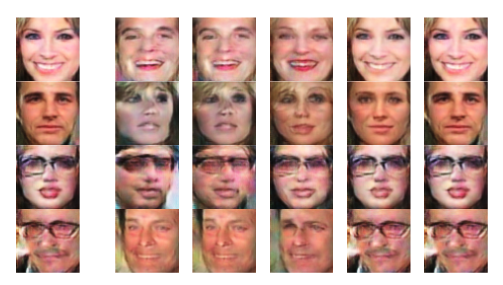}}
  \caption{Recovery (from generated images) process visualization for (a) MNIST dataset, and (b) CelebA dataset. For both figures from left to right columns showing: $G(\bm{z},~\bm{y})$ from true $\bm{z}$ and $\bm{y}$, $G(\bm{z_p,~\bm{y_p}})$ from probe $\bm{z_p}$ and $\bm{y_p}$ after initialization, $10$ iterations, $100$ iterations, $1,000$ iterations and $10,000$ iterations.}
  \label{fig:g_v}
\end{figure}

The recovery process from initialized probe vector $\bm{z_p}$ and $\bm{y_p}$ towards the true vector $\bm{z}$ and $\bm{y}$ is visualized by $G(\bm{z_p},~\bm{y_p})$ generated from them, during the iterations of optimization process. In Fig.~\ref{fig:mnist_g_v} and~\ref{fig:celeb_g_v}, $G(\bm{z_p,~\bm{y_p}})$ after initialization, 10 iterations, 100 iterations, 1,000 iterations and 10,000 iterations are shown, together with the generated image $G(\bm{z},~\bm{y})$ from true $\bm{z}$ and $\bm{y}$.
The true conditional vectors $\bm{y}$ in Fig.~\ref{fig:mnist_g_v} and~\ref{fig:celeb_g_v} are $[3, 6, 7, 9, 5, 0, 2, 8, 1, 4]$ and $[0,1,2,3]$ in one-hot encoding.
We can see that when initialized, the $G(\bm{z_p},~\bm{y_p})$ looks completely different from the $G(\bm{z},~\bm{y})$. The initialized images are also of visually bad quality due to the fact that $\bm{y_p}$ is initialized as a zero vector instead of a valid one-hot encoded vector. As the number of iterations increases, $G(\bm{z_p},~\bm{y_p})$ becomes more and more visually similar to $G(\bm{z},~\bm{y})$. After $10k$ iterations, $G(\bm{z_p},~\bm{y_p})$ is visually indistinguishable from $G(\bm{z},~\bm{y})$.

\begin{figure}[h!]
\centering
\subfloat[\,\label{fig:m_g_rloss}]{\includegraphics[width=0.34\columnwidth]{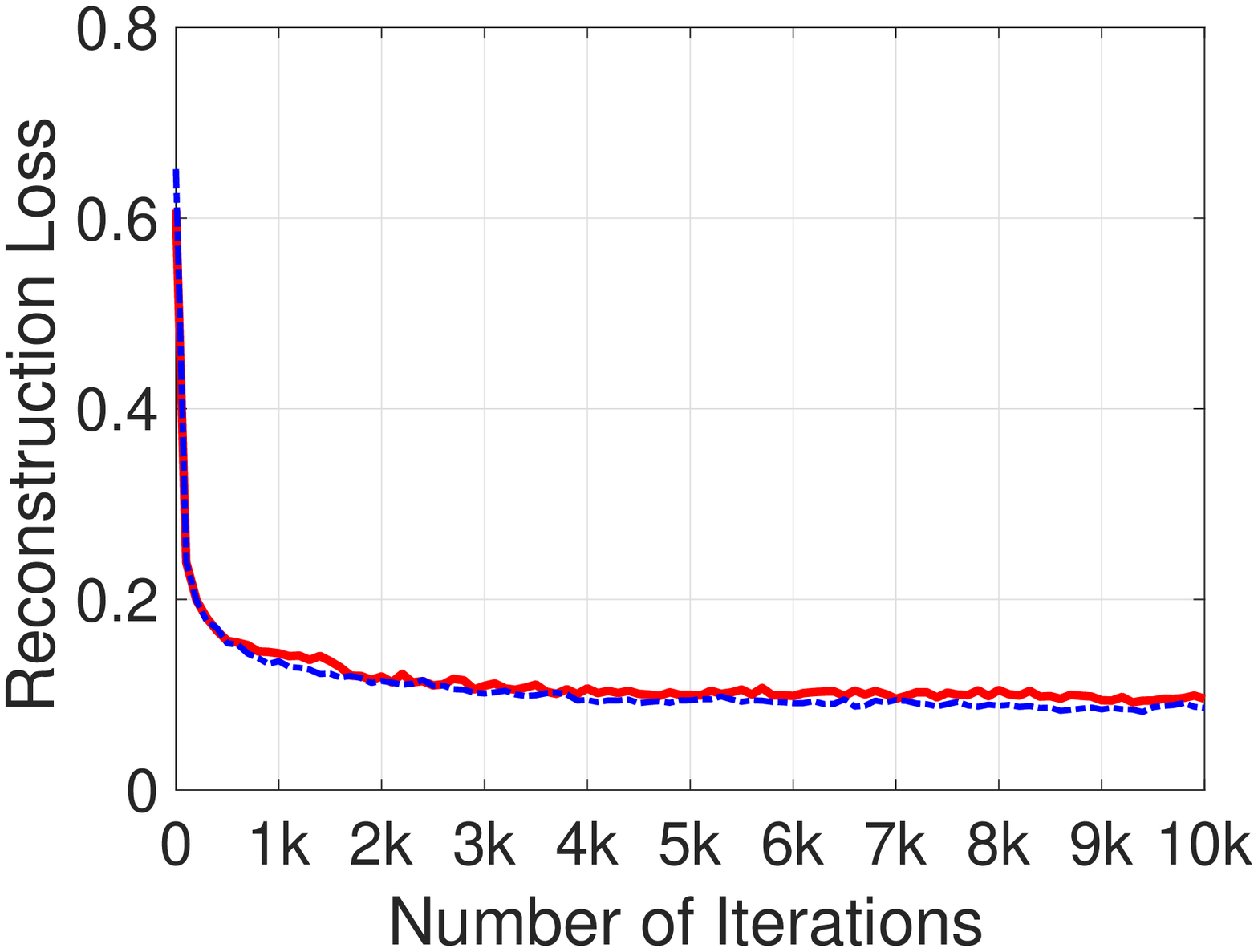}}
\subfloat[\,\label{fig:m_g_zloss}]{\includegraphics[width=0.34\columnwidth]{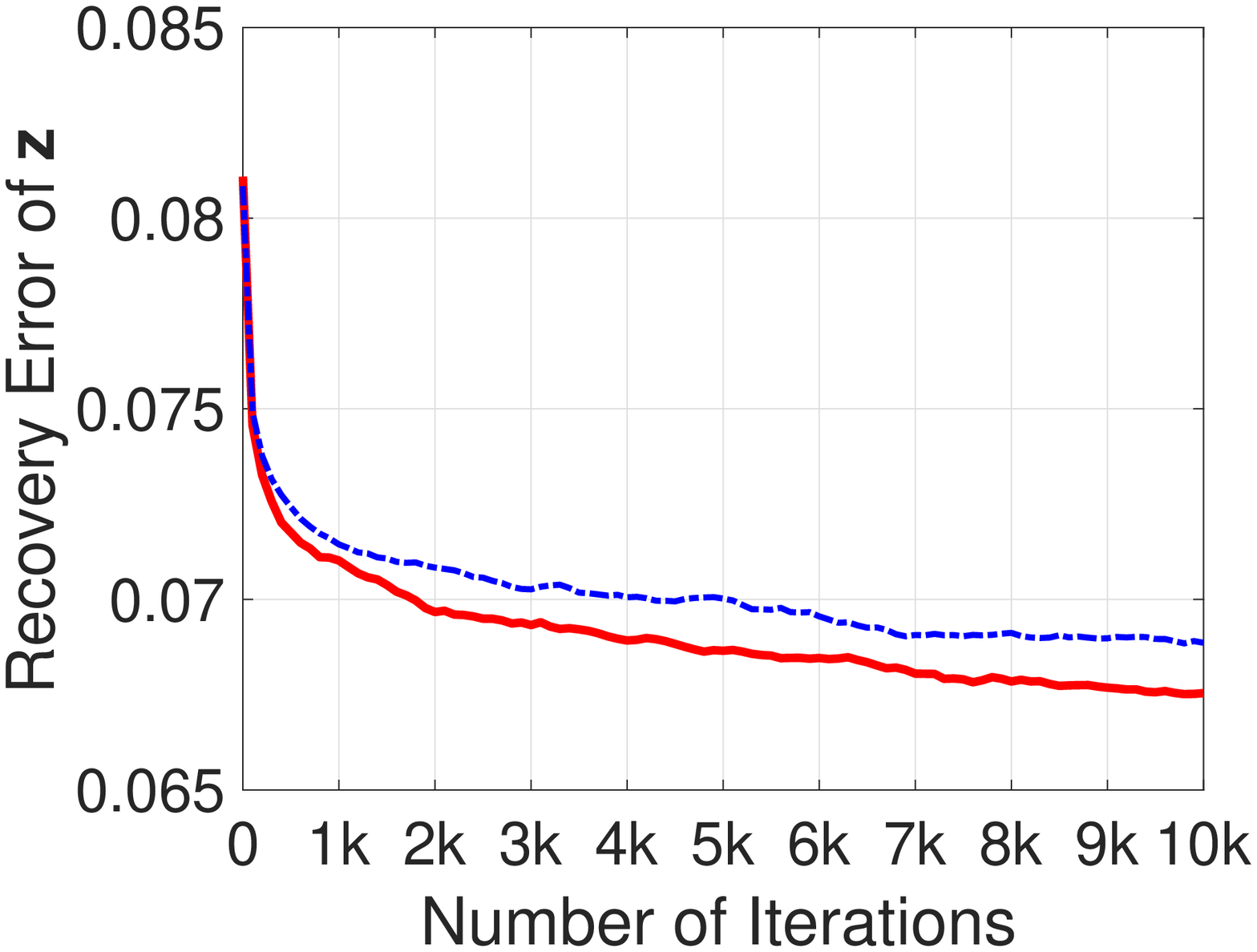}}
\subfloat[\,\label{fig:m_g_yacc}]{\includegraphics[width=0.34\columnwidth]{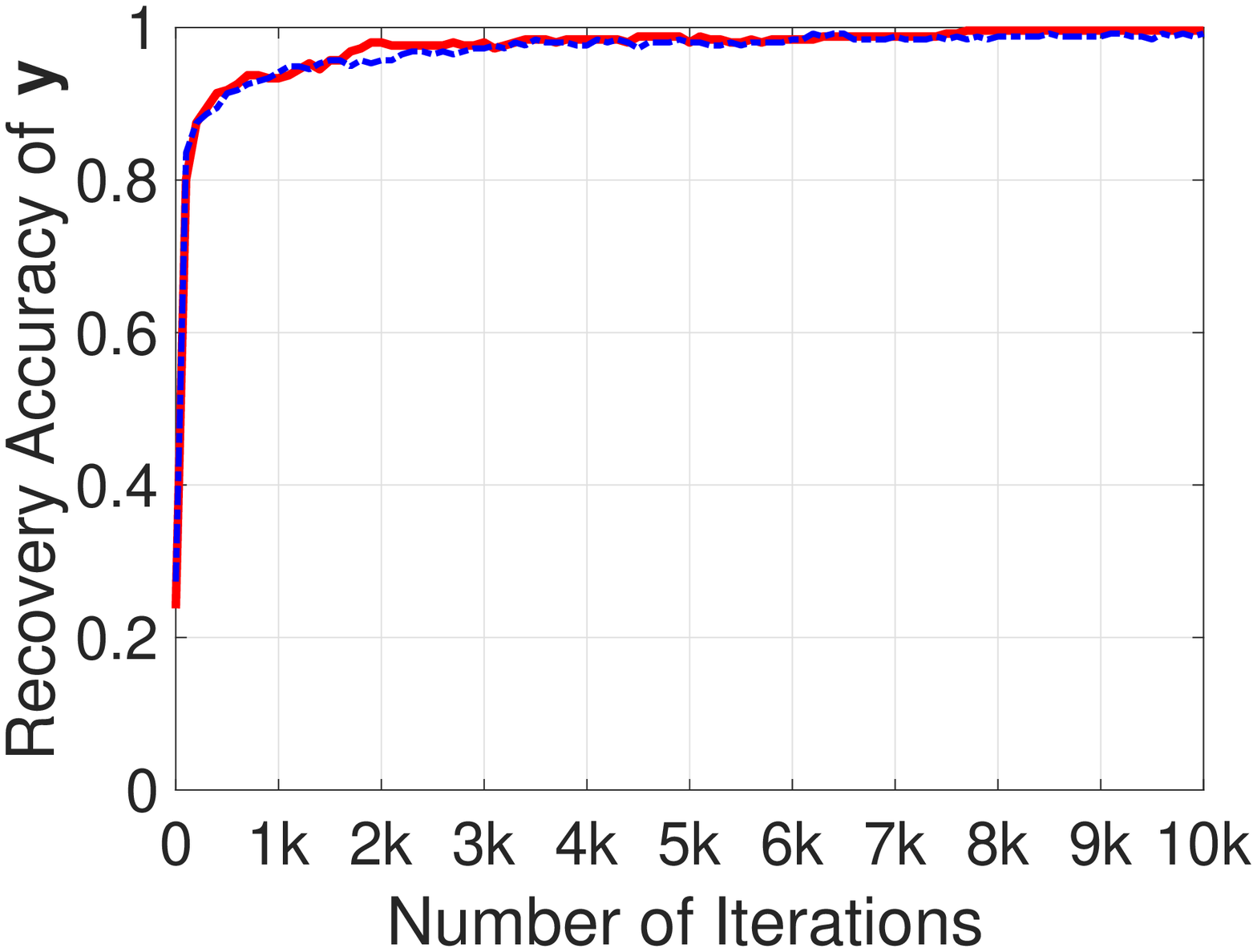}}
\caption{In MNIST dataset: (a) reconstruction loss; (b) recovery error of $\bm{z}$; (c) recovery accuracy of $\bm{y}$. The red solid line is with regularization while the blue dashed line is without regularization.}
\label{fig:m_g}
\end{figure}

\begin{figure}[h!]
\centering
\subfloat[\,\label{fig:c_g_rloss}]{\includegraphics[width=0.34\columnwidth]{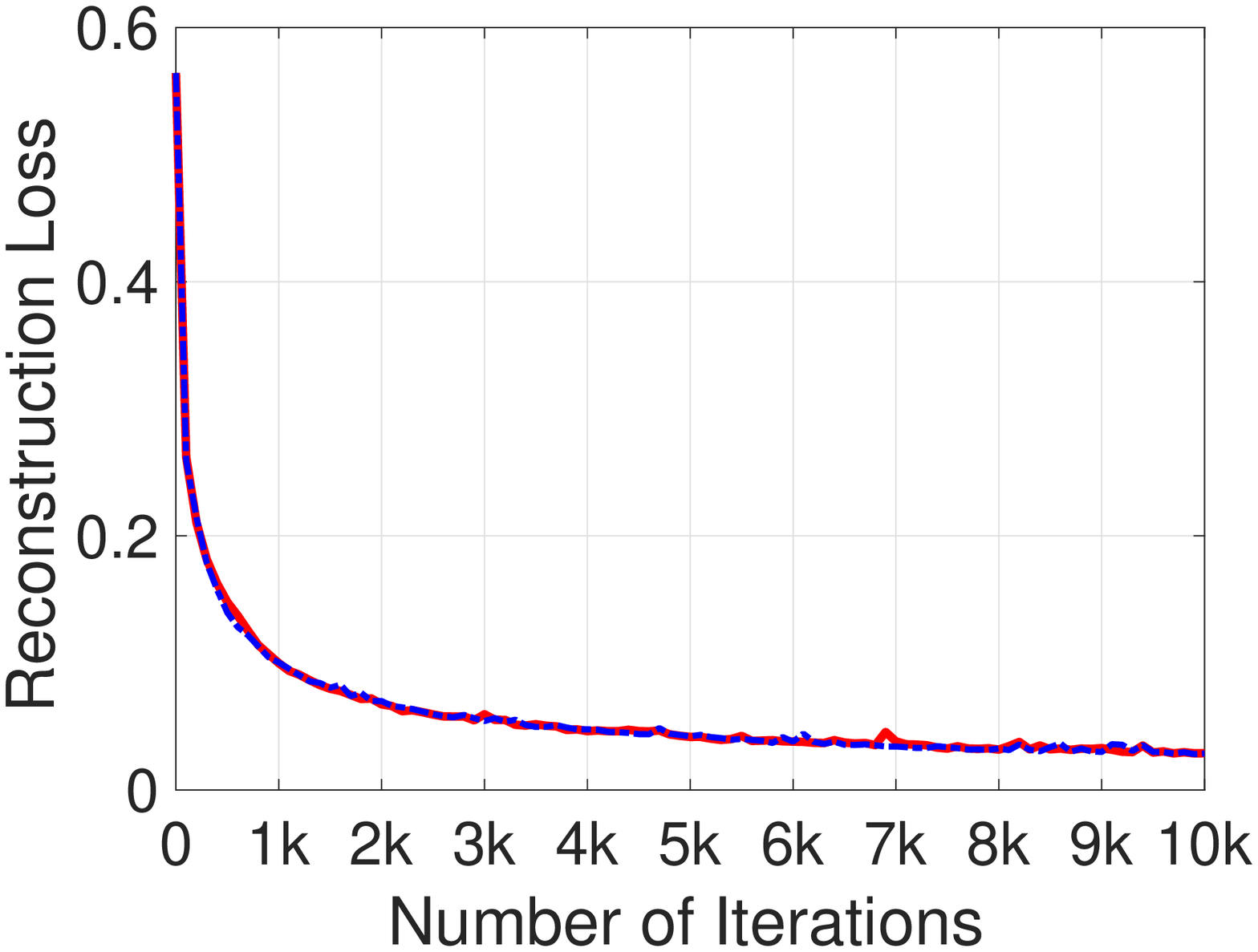}}
\subfloat[\,\label{fig:c_g_zloss}]{\includegraphics[width=0.34\columnwidth]{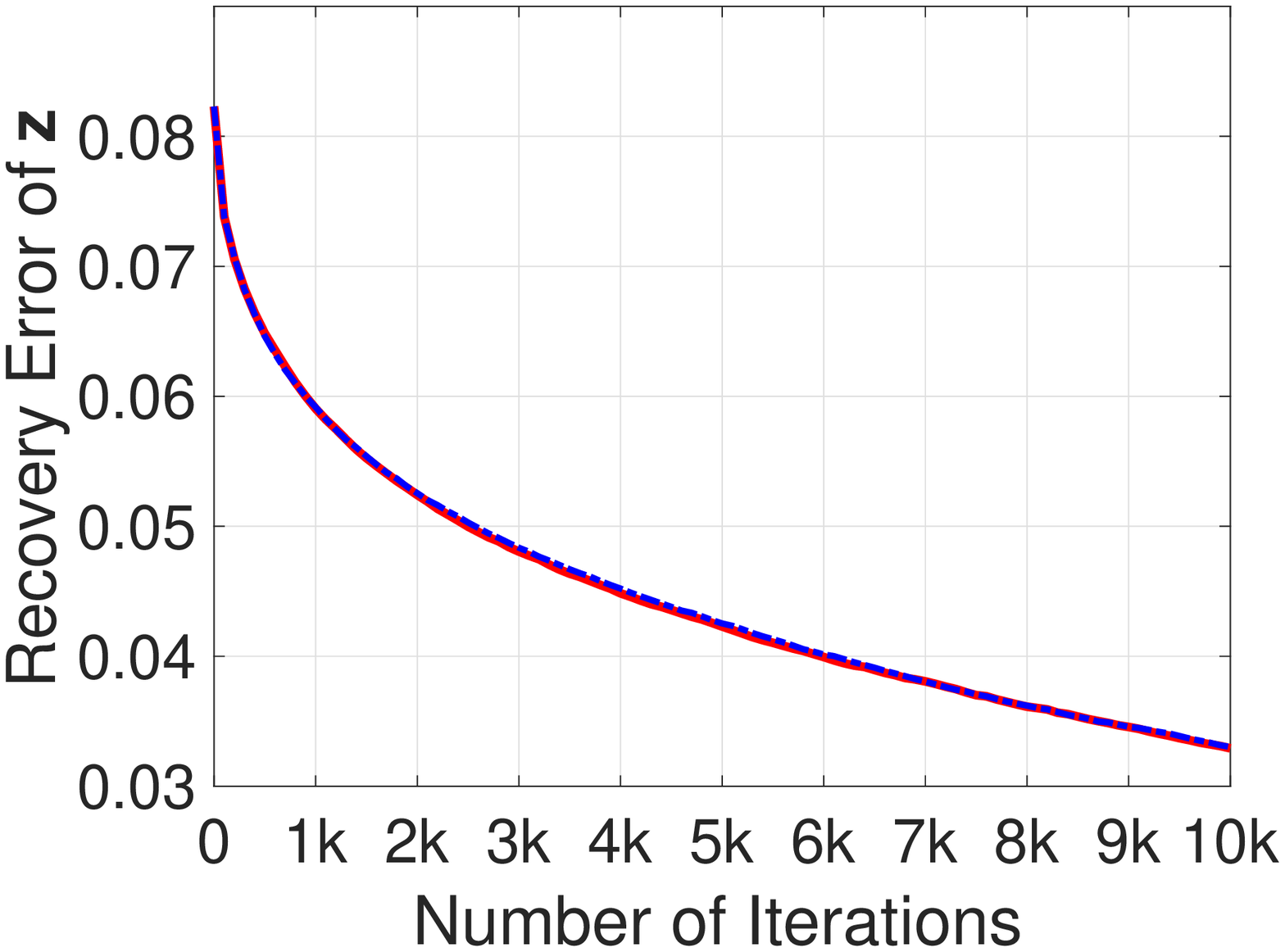}}
\subfloat[\,\label{fig:c_g_yacc}]{\includegraphics[width=0.34\columnwidth]{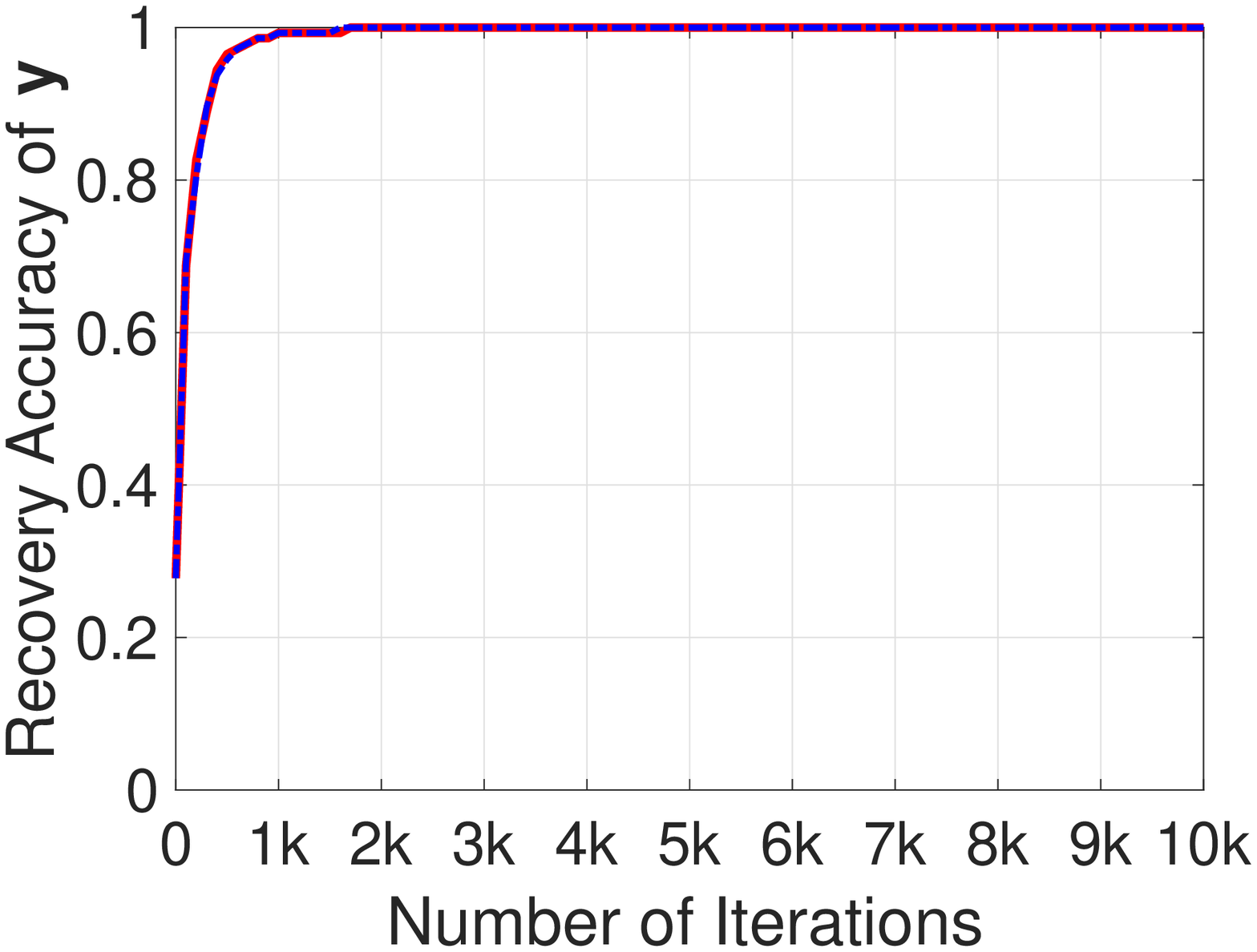}}
\caption{In CelebA dataset: (a) reconstruction loss; (b) recovery error of $\bm{z}$; (c) recovery accuracy of $\bm{y}$. The red solid line is with regularization while the blue dashed line is without regularization.}
\label{fig:c_g}
\end{figure}

Reconstruction loss is defined as the mean squared error per pixel of the reconstructed image, with value scaled to $[-1,~1]$. A successful recovery of $\bm{z}$ and $\bm{y}$ should generate a small reconstruction loss. The recovery error of $\bm{z}$ is defined as the Euclidean distance between the true $\bm{z}$ and the probe $\bm{z_p}$. The recovery accuracy of $\bm{y}$ is calculated after taking $\argmax(\bm{y_p})$, i.e., the index of the maximum value in one-hot encoded vector is reported as final recovered conditional label. The first $10k$ iterations of one batch are plotted for these values in Fig.~\ref{fig:m_g} and~\ref{fig:c_g} with (Eq.~\ref{eq:objective-function-cond-reg}) and without regularization (Eq.~\ref{eq:objective-function-cond}). We can see that image reconstruction loss and recovery error of $\bm{z}$ decrease rapidly in the first few hundreds of iterations, and continue to decrease slowly afterwards. The accuracy of recovery conditional vector $\bm{y}$ approaches $100\%$ rather quickly and steadily. In MNIST dataset, the reconstruction loss with and without the regularization are similar; the recovery error of $\bm{z}$ is lower when the regularization is applied; the recovery accuracy of $\bm{y}$ is marginally higher with regularization. In CelebA dataset, they are almost identical. This shows that \emph{for generated images}, the recovery of conditional vector can be achieved with high accuracy using a gradient based method with or without extra regularization.

\subsection{Recovery from Real Images}\label{sec:recover-real-images}
We further apply the recovery operation on real images. It is interesting to study the effect of conditional vector recovery, when the images are real and not generated by the generator. The same experiments as in previous subsection are repeated this time for real images.

\begin{figure}[h!]
  \centering
  \subfloat[MNIST\label{fig:mnist_r_v}]{\includegraphics[width=0.34\columnwidth]{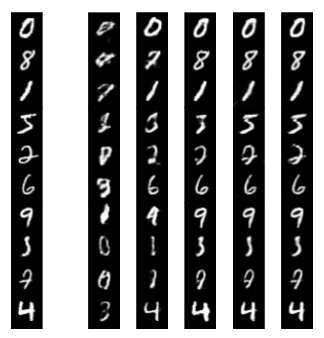}}~~~~
  \subfloat[CelebA\label{fig:celeb_r_v}]{\includegraphics[width=0.64\columnwidth]{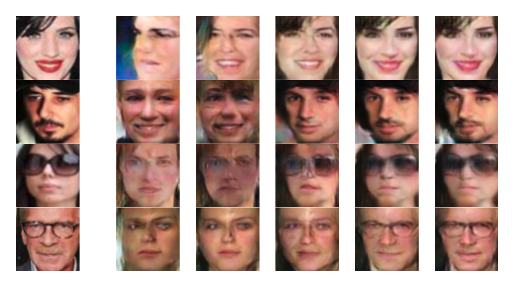}}
  \caption{Recovery (from real images) process visualization for (a) MNIST dataset, and (b) CelebA dataset. For both figures from left to right columns showing: real images, $G(\bm{z_p,~\bm{y_p}})$ from probe $\bm{z_p}$ and $\bm{y_p}$ after initialization, 10 iterations, 100 iterations, 1,000 iterations and 10,000 iterations.}
  \label{fig:r_v}
\end{figure}

Again, we first examine the recovery process visually through reconstructed images. In Fig.~\ref{fig:mnist_r_v} and~\ref{fig:celeb_r_v}, the real images and the process of approaching these real images with generated images are illustrated. The images transfer from the randomly initialized ones to the ones that are visually similar to the target images. However, it can still be observed that they are not exactly the same even visually. This is more obvious in CelebA dataset. In Fig.~\ref{fig:mnist_r_v}, an example of incorrectly recovered conditional vector is shown, which is the $5^{th}$ row. The true $\bm{y}$ is of label ``$2$'', while the recovered $\bm{y_p}$ is of label ``$7$''. It is interesting to observe how the network manages to produce an image that is very close to digit ``$2$'' given the condition ``$7$''. Actually, the fact that the reconstruction loss can be low even with the incorrect $\bm{z}$ and $\bm{y}$ is the main challenge we encountered.

\begin{figure}[h!]
\centering
\subfloat[\,\label{fig:m_r_rloss}]{\includegraphics[width=0.415\columnwidth]{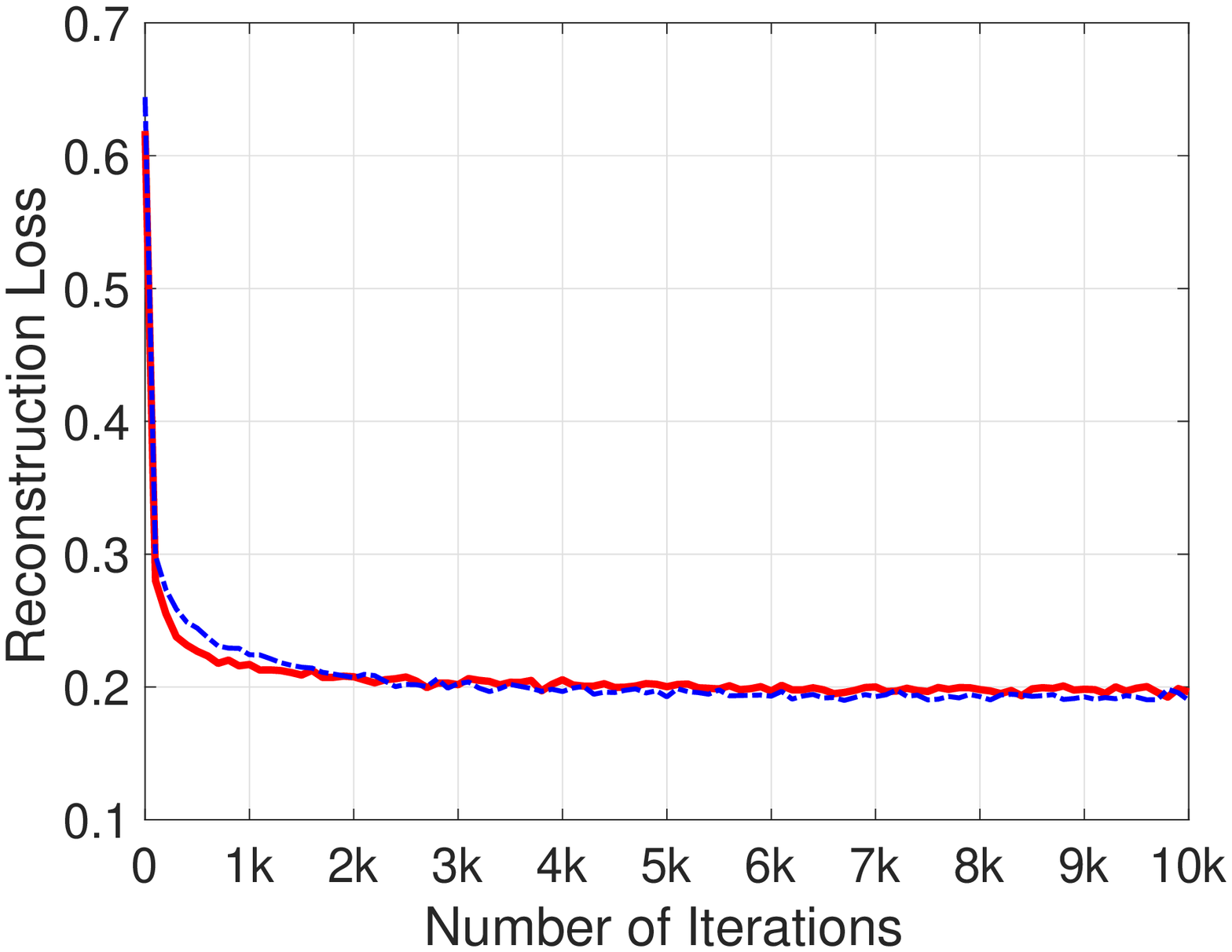}}~~~
\subfloat[\,\label{fig:m_r_yacc}]{\includegraphics[width=0.415\columnwidth]{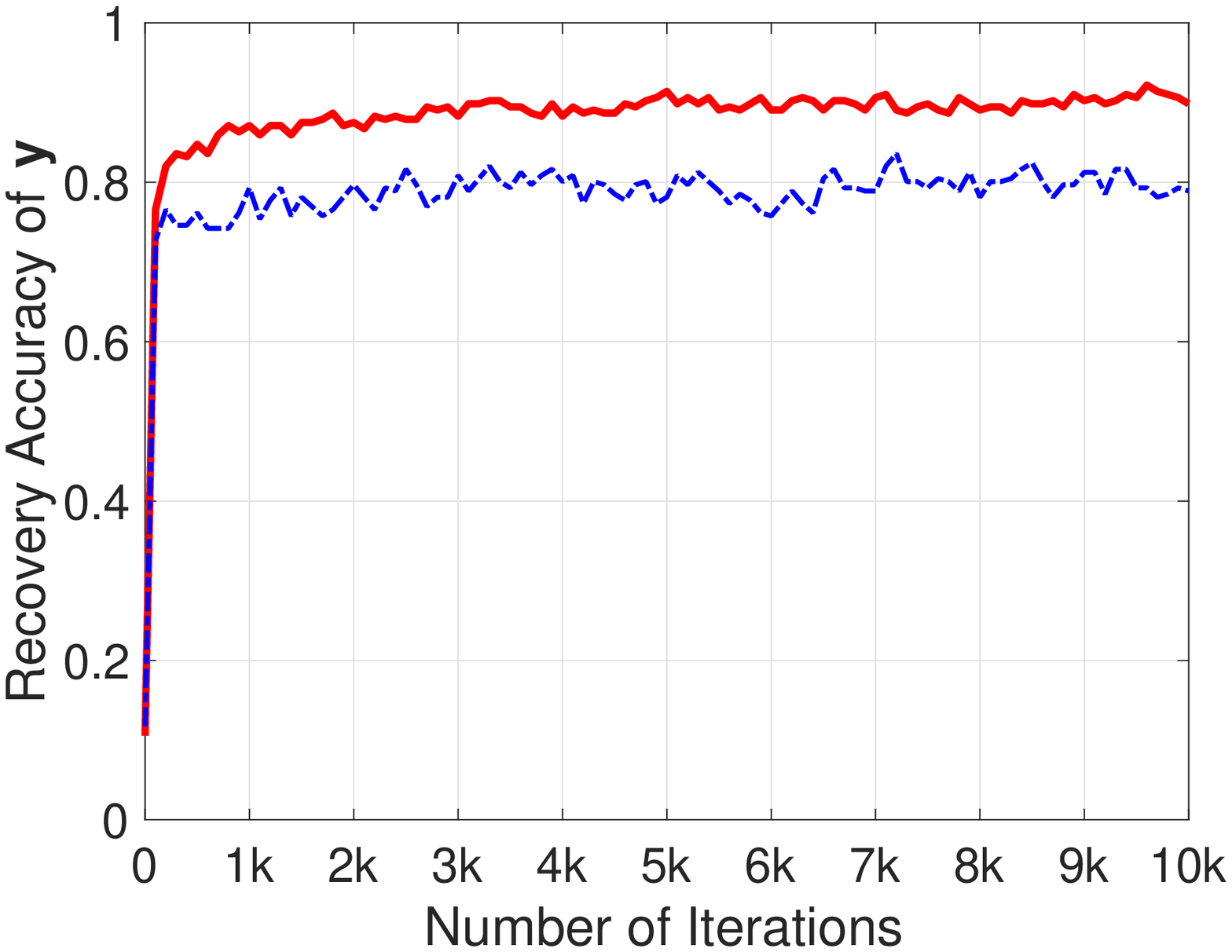}}
\caption{In MNIST dataset: (a) reconstruction loss; (b) recovery accuracy of $\bm{y}$. The red solid line is with regularization while the blue dashed line is without regularization.}
\label{fig:m_r}
\end{figure}

\begin{figure}[h!]
\centering
\subfloat[\,\label{fig:c_r_rloss}]{\includegraphics[width=0.415\columnwidth]{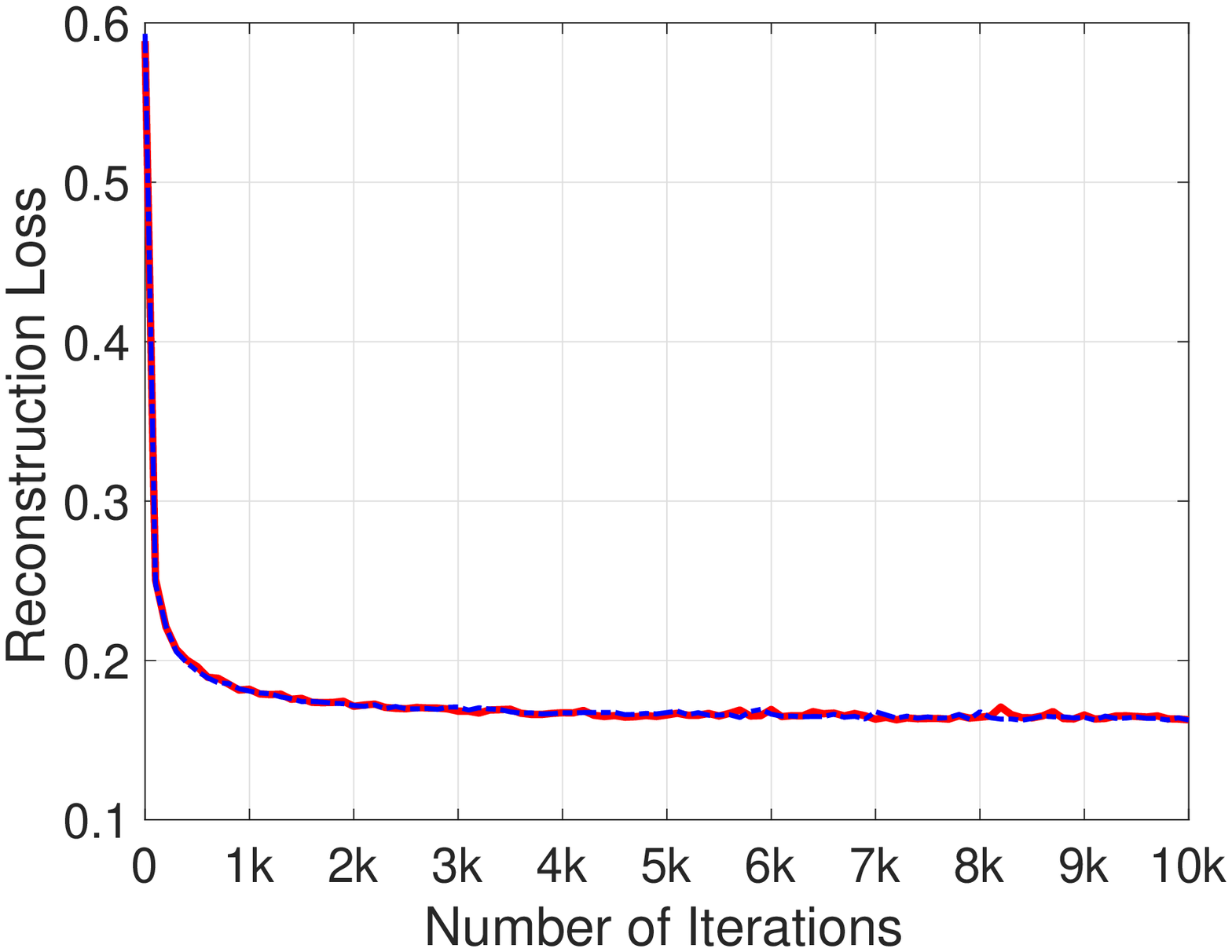}}~~~
\subfloat[\,\label{fig:c_r_yacc}]{\includegraphics[width=0.415\columnwidth]{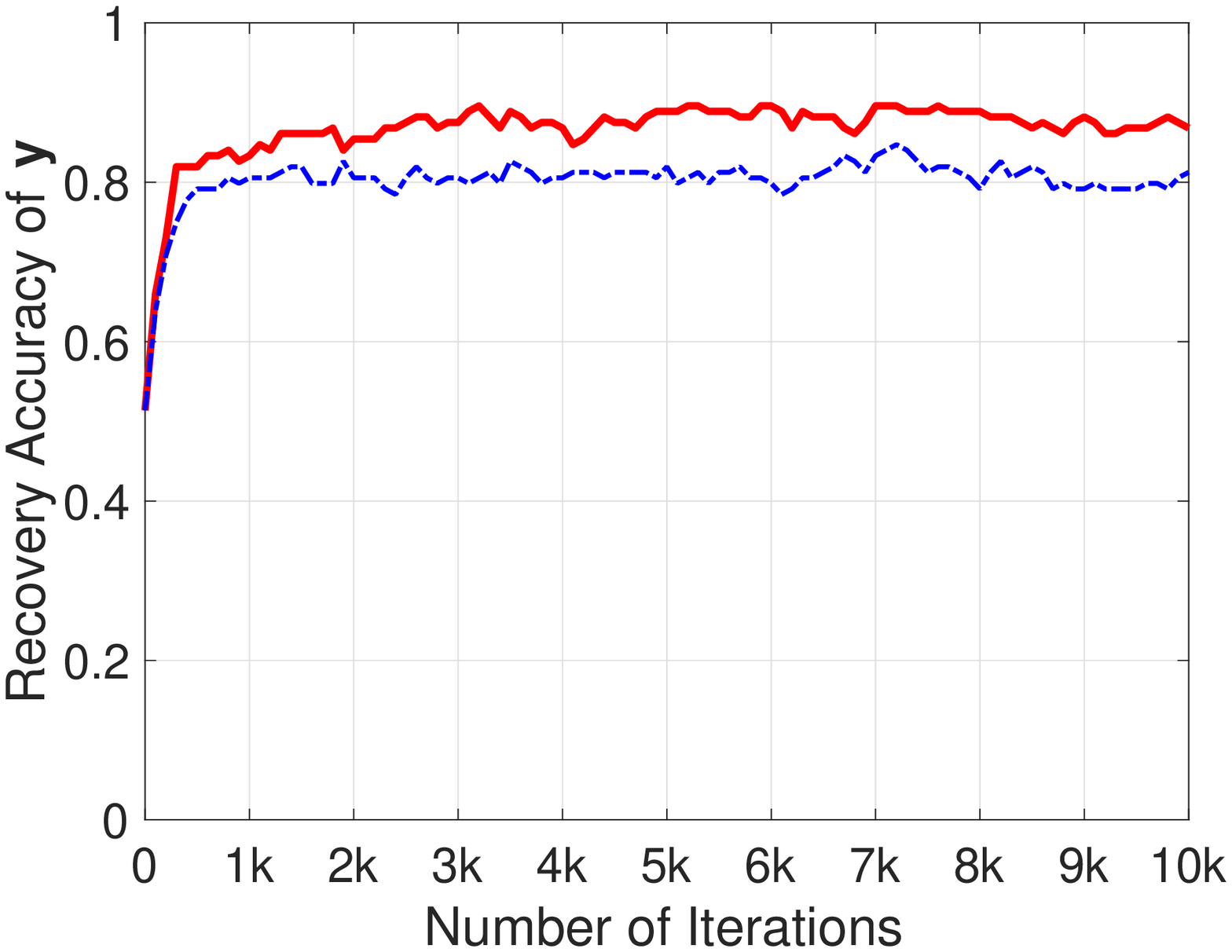}}
\caption{In CelebA dataset: (a) reconstruction loss; (b) recovery accuracy of $\bm{y}$. The red solid line is with regularization while the blue dashed line is without regularization.}
\label{fig:c_r}
\end{figure}

The reconstruct loss and recovery accuracy of $\bm{y}$ for the first $10k$ iterations of one batch are plotted in Fig.~\ref{fig:m_r} and~\ref{fig:c_r}. Notice this time there isn't a true $\bm{z}$ of a real image for us to compute recovery error of $\bm{z}$. Compared with recovery from generated images, the reconstruction loss is greater and the recovery accuracy of $\bm{y}$ is lower when recovering from real images. Especially the recovery of $\bm{y}$ becomes less stable, frequently toggling between two possible values back and forth for some images. Very importantly, for real images, there exists a quite significant gap ($5\%-10\%$) between with and without regularization for the recovery accuracy of $\bm{y}$, in both MNIST and CelebA datasets, showing that the regularization improves the recovery accuracy of conditional vector for real images.

\begin{table*}[th!]
\caption{Results after $100k$ iterations.}
\label{tab:1}
\centering
\bgroup
\def\arraystretch{1.5}
\begin{tabular}{C{2.0cm} C{2.7cm} C{3.3cm} C{3.8cm}}
\hlineB{3}
\textbf{Dataset} & \textbf{Recovered From} & \textbf{Reconstruction Loss} & \textbf{Recovery Accuracy of $\bm{y}$}\\ \hlineB{2}
MNIST &  $G(\bm{z},~\bm{y})$ & $0.0176~(0.6105)$ & $0.9985$ \\
MNIST & $I_{real}$ & $0.1382~(0.6177)$ & $0.9033$\\
CelebA & $G(\bm{z},~\bm{y})$ & $0.0042~(0.5781)$ & $1.0000$ \\
CelebA & $I_{real}$ & $0.1563~(0.5780)$ & $0.8359$ \\
\hlineB{2}
\end{tabular}
\egroup
\end{table*}
\subsection{Converged Results}
The optimization is considered converged after $100k$ iterations (most of the time it takes less iterations) from empirically observation.
The results after running the optimization for $100k$ iterations with the proposed method are listed in Table~\ref{tab:1}. $G(\bm{z},~\bm{y})$ represents the generated images and $I_{real}$ represents the real images. In \emph{Reconstruction Loss} column, numbers in the brackets represent the initial losses. It shows that it is easy to recover the conditional vectors from generated images, and the reconstruction loss can be very low. On the other hand, for real images, the recovery is not always successful and the original images can not always be reconstructed exactly, which means it is often impossible to generate certain real images. This can always happen when the underlying data distribution modeled by the generator is not perfect.

\section{Discussion}
We noticed that a better recovered $\bm{z_p}$ does not necessarily result in better reconstruction loss (Fig.~\ref{fig:m_g_rloss} and~\ref{fig:m_g_zloss}). Also a much better recovery of $\bm{y_p}$ does not translate to equal amount of advantage in reconstruction loss (Fig.~\ref{fig:m_r} and~\ref{fig:c_r}). One possible explanation is that, for one image, there are multiple (potentially infinite) combinations and values of $\bm{z}$ and $\bm{y}$ from which the it can be generated. Another point could be that the reconstruction loss is not the most appropriate evaluation metric.
The objective function used in this work is based on reconstruction loss, which evaluates per pixel differences in image domain. It would be interesting to see if other losses, for example, mean squared error of discriminative CNN features, can produce better gradient thus lead to better results.

While the recovery of conditional vectors has similar performance across the two datasets, the recovery of latent vectors differs a lot. The recovery error of $\bm{z}$ reduces much slower on MNIST than CelebA dataset. It is possible that $\bm{z}$ is utilized to a ``greater extent'' in CelebA because of much more complex content compared with MNIST (color faces vs. gray scale digits).
We suspect that in MNIST dataset, some $\bm{z}$ mapped to the same image, or some dimensions of $\bm{z}$ become basically irrelevant. More investigation of how different dimensions of $\bm{z}$ and $\bm{y}$ impact the recovery could be worthy. Again, the cGAN used in this experiment is a simple DCGAN structure, with more recent advancement in the training of GAN such as~\cite{{arjovsky2017towards},{che2016mode},{warde2016improving}}, the performance of recover is expected to improve.

The ability to access the latent and conditional vector of a given generated image from a cGAN could potentially be used for tasks such as debugging and diagnosis of the network. Even though we could not calculate the recovery error for $\bm{z}$ for real images, we do get consistent $\bm{z_p}$ for the same image. It remains interesting to see if this could be applied to detect adversarial attacks. An adversarial image could have different behaviour in terms of $\bm{z_p}$ and $\bm{y_p}$ when being recovered.

\section{Conclusion}
In this work, we show that it is possible to recover the latent vector as well as the conditional vector from a conditional generative adversarial network. The approach could potentially enable a wide spectrum of applications ranging from image manipulation for entertaining purposes to diagnosis of the neural networks for security purposes. The method minimizes a regularized reconstruction loss using projected gradient descent and stochastic clipping. The regularizer is designed for conditional vector being discrete labels. The recovery method is evaluated on two public datasets for both generated images and real images. We see that the conditional vector can be recovered with high accuracy from generated images, and to a lesser extent from real images. The result is promising, and how to close the gap between recovery from generated images and real images will be our future direction.

%

\bibliography{rgan}

\begin{thebibliography}{10}

\bibitem{arjovsky2017towards}
Martin Arjovsky and L{\'e}on Bottou.
\newblock Towards principled methods for training generative adversarial
  networks.
\newblock In {\em International Conference on Learning Representations (ICLR)},
  2017.

\bibitem{che2016mode}
Tong Che, Yanran Li, Athul~Paul Jacob, Yoshua Bengio, and Wenjie Li.
\newblock Mode regularized generative adversarial networks.
\newblock In {\em International Conference on Learning Representations (ICLR)},
  2017.

\bibitem{creswell2016inverting}
Antonia Creswell and Anil~Anthony Bharath.
\newblock Inverting the generator of a generative adversarial network.
\newblock In {\em Workshop on Adversarial Training, NIPS}, 2016.

\bibitem{donahue2016adversarial}
Jeff Donahue, Philipp Kr{\"a}henb{\"u}hl, and Trevor Darrell.
\newblock Adversarial feature learning.
\newblock In {\em International Conference on Learning Representations (ICLR)},
  2017.

\bibitem{dumoulin2016adversarially}
Vincent Dumoulin, Ishmael Belghazi, Ben Poole, Alex Lamb, Martin Arjovsky,
  Olivier Mastropietro, and Aaron Courville.
\newblock Adversarially learned inference.
\newblock In {\em International Conference on Learning Representations (ICLR)},
  2017.

\bibitem{gauthier2014conditional}
Jon Gauthier.
\newblock Conditional generative adversarial nets for convolutional face
  generation.
\newblock {\em Class Project for Stanford CS231N}, 2014.

\bibitem{goodfellow2016nips}
Ian Goodfellow.
\newblock Nips 2016 tutorial: Generative adversarial networks.
\newblock {\em arXiv preprint arXiv:1701.00160}, 2016.

\bibitem{goodfellow2014generative}
Ian Goodfellow, Jean Pouget-Abadie, Mehdi Mirza, Bing Xu, David Warde-Farley,
  Sherjil Ozair, Aaron Courville, and Yoshua Bengio.
\newblock Generative adversarial nets.
\newblock In {\em Advances in neural information processing systems (NIPS)},
  pages 2672--2680, 2014.

\bibitem{isola2016image}
Phillip Isola, Jun-Yan Zhu, Tinghui Zhou, and Alexei~A Efros.
\newblock Image-to-image translation with conditional adversarial networks.
\newblock In {\em Proceedings of the IEEE conference on Computer Vision and
  Pattern Recognition (CVPR)}, 2017.

\bibitem{carpedm20}
Taehoon Kim.
\newblock A tensorflow implementation of ``deep convolutional generative
  adversarial networks''.
\newblock \url{https://github.com/carpedm20/DCGAN-tensorflow}, 2016.

\bibitem{lecun1998gradient}
Yann LeCun, L{\'e}on Bottou, Yoshua Bengio, and Patrick Haffner.
\newblock Gradient-based learning applied to document recognition.
\newblock {\em Proceedings of the IEEE}, 86(11):2278--2324, 1998.

\bibitem{li2016precomputed}
Chuan Li and Michael Wand.
\newblock Precomputed real-time texture synthesis with markovian generative
  adversarial networks.
\newblock In {\em European Conference on Computer Vision (ECCV)}, pages
  702--716. Springer, 2016.

\bibitem{lipton2017precise}
Zachary~C Lipton and Subarna Tripathi.
\newblock Precise recovery of latent vectors from generative adversarial
  networks.
\newblock In {\em International Conference on Learning Representations (ICLR)
  Workshop Track}, 2017.

\bibitem{liu2015faceattributes}
Ziwei Liu, Ping Luo, Xiaogang Wang, and Xiaoou Tang.
\newblock Deep learning face attributes in the wild.
\newblock In {\em Proceedings of International Conference on Computer Vision
  (ICCV)}, 2015.

\bibitem{mahendran2015understanding}
Aravindh Mahendran and Andrea Vedaldi.
\newblock Understanding deep image representations by inverting them.
\newblock In {\em Proceedings of the IEEE conference on Computer Vision and
  Pattern Recognition (CVPR)}, pages 5188--5196, 2015.

\bibitem{mathieu2015deep}
Michael Mathieu, Camille Couprie, and Yann LeCun.
\newblock Deep multi-scale video prediction beyond mean square error.
\newblock In {\em International Conference on Learning Representations (ICLR)},
  2016.

\bibitem{mirza2014conditional}
Mehdi Mirza and Simon Osindero.
\newblock Conditional generative adversarial nets.
\newblock {\em arXiv preprint arXiv:1411.1784}, 2014.

\bibitem{pathak2016context}
Deepak Pathak, Philipp Krahenbuhl, Jeff Donahue, Trevor Darrell, and Alexei~A
  Efros.
\newblock Context encoders: Feature learning by inpainting.
\newblock In {\em Proceedings of the IEEE Conference on Computer Vision and
  Pattern Recognition (CVPR)}, pages 2536--2544, 2016.

\bibitem{radford2015unsupervised}
Alec Radford, Luke Metz, and Soumith Chintala.
\newblock Unsupervised representation learning with deep convolutional
  generative adversarial networks.
\newblock In {\em International Conference on Learning Representations (ICLR)
  Workshop Track}, 2016.

\bibitem{reed2016generative}
Scott Reed, Zeynep Akata, Xinchen Yan, Lajanugen Logeswaran, Bernt Schiele, and
  Honglak Lee.
\newblock Generative adversarial text to image synthesis.
\newblock In {\em International Conference on Machine Learning (ICML)}, 2016.

\bibitem{salimans2016improved}
Tim Salimans, Ian Goodfellow, Wojciech Zaremba, Vicki Cheung, Alec Radford, and
  Xi~Chen.
\newblock Improved techniques for training gans.
\newblock In {\em Advances in Neural Information Processing Systems (NIPS)},
  pages 2234--2242, 2016.

\bibitem{warde2016improving}
David Warde-Farley and Yoshua Bengio.
\newblock Improving generative adversarial networks with denoising feature
  matching.
\newblock In {\em International Conference on Learning Representations (ICLR)},
  2017.

\bibitem{zhu2016generative}
Jun-Yan Zhu, Philipp Kr{\"a}henb{\"u}hl, Eli Shechtman, and Alexei~A Efros.
\newblock Generative visual manipulation on the natural image manifold.
\newblock In {\em European Conference on Computer Vision (ECCV)}, pages
  597--613. Springer, 2016.

\bibitem{CycleGAN2017}
Jun-Yan Zhu, Taesung Park, Phillip Isola, and Alexei~A Efros.
\newblock Unpaired image-to-image translation using cycle-consistent
  adversarial networks.
\newblock In {\em IEEE International Conference on Computer Vision (ICCV)},
  2017.

\end{thebibliography}
\bibliographystyle{plain}

\end{document}